\documentclass{bmvc2k}

\usepackage{bmvc2k_natbib}
\usepackage{multirow}
\usepackage{booktabs}       
\usepackage{dsfont}

\title{Image Captioning with Unseen Objects}

\addauthor{Berkan Demirel}{berkan.demirel@metu.edu.tr}{1}
\addauthor{Ramazan Gokberk Cinbis}{gcinbis@ceng.metu.edu.tr}{2}
\addauthor{Nazli Ikizler-Cinbis}{nazli@cs.hacettepe.edu.tr}{3}

\addinstitution{
 HAVELSAN Inc. \& Middle East Technical University\\
 Ankara, Turkey
}
\addinstitution{
 Middle East Technical University\\
 Ankara, Turkey
}
\addinstitution{
 Hacettepe University\\
 Ankara, Turkey
}

\runninghead{Demirel, Cinbis, Ikizler-Cinbis}{Image Captioning with Unseen Objects}

\def\eg{\emph{e.g}\bmvaOneDot}

\def\etal{\emph{et al}\bmvaOneDot}
 \input{macros}

\IfFileExists{"../customcolors.tex"}{\input{"../customcolors.tex"}}{} 

\begin{document}

\maketitle

\begin{abstract}
Image caption generation is a long standing and challenging problem at the intersection of computer vision and natural language processing. A number of recently proposed approaches utilize a fully supervised object recognition model within the captioning approach. Such models, however, tend to generate sentences which only consist of objects predicted by the recognition models, excluding instances of the classes without labelled training examples. In this paper, we propose a new challenging scenario that targets the image captioning problem in a fully zero-shot learning setting, where the goal is to be able to generate captions of test images containing objects that are not seen during training. The proposed approach jointly uses a novel zero-shot object detection model and a template-based sentence generator. Our experiments show promising results on the COCO dataset.
\end{abstract}

\section{Introduction}
\label{sec:intro}

Image captioning, the problem of generating a concise textual summary of a given image, is one of the most actively
studied problems standing at the intersection of computer vision and natural language processing.  Following the success
of deep learning based object detection approaches~\cite{redmon2016yolo9000, redmon2016you, girshick2015fast,
sermanet2014overfeat, bell2016inside, ren2015faster, lin2017feature, liu2016ssd, yan2014fastest,lin2018focal,
law2018cornernet}, there have been a recent interest in generating visually grounded image captions by constructing
captioning models that operate on the outputs of supervised object detectors~\cite{kulkarni2013babytalk, anne2016deep,
yin2017obj2text, lu2018neural}. However, the success of such approaches are inherently limited by the set of
classes provided in the detector training dataset, which is typically too small to construct a visually comprehensive model. 
Therefore, such models are arguably unsuitable for captioning in uncontrolled images, which are
likely to contain object classes unseen at training time.

In the context of image classification, {\em zero-shot learning} (ZSL) has emerged as a promising alternative towards
overcoming the practical limits in collecting labelled image datasets and constructing image
classifiers with very large object vocabularies. ZSL approaches typically tackle the problem of transferring visual and/or
semantic knowledge from the {\em seen} classes (training classes) to {\em unseen classes} (novel classes appearing at
test time) by using a variety of techniques and information sources, such as class
attributes~\cite{akata2013label,jayaraman2014zero}, class hierarchies~\cite{rohrbach2011evaluating,akata2015evaluation},
attribute-to-class name mappings~\cite{demirel2017attributes2classname}, label relation
graphs~\cite{deng2014large}. While ZSL remains as an unsolved problem, significant progress have been made in recent
years in zero-shot image classification~\cite{xian2018zero}. Very recently, a few methods have also been proposed for
the {\em zero-shot object detection} (ZSD) problem~\cite{demirel2018zero, rahman2018zero, bansal2018zero}.

In a similar manner, {\em zero-shot image captioning} (ZSC), where the goal is to create
captions of images containing objects classes unseen at the training time, have the promise of overcoming the data
collection bottleneck in image captioning. However, we observe that there is no benchmark directly tailored to study captioning in a true
zero-shot setting, and, there is no prior work explicitly towards addressing true ZSC.
While there are few very recent works on ZSC~\cite{lu2018neural, yao2017incorporating}, all these methods study the problem purely 
as a language modeling problem and presume the availability of a pre-trained
fully-supervised object detector over all object classes of interest, to the best of our knowledge.

\begin{figure*}
\begin{center}
\includegraphics[width=\textwidth]{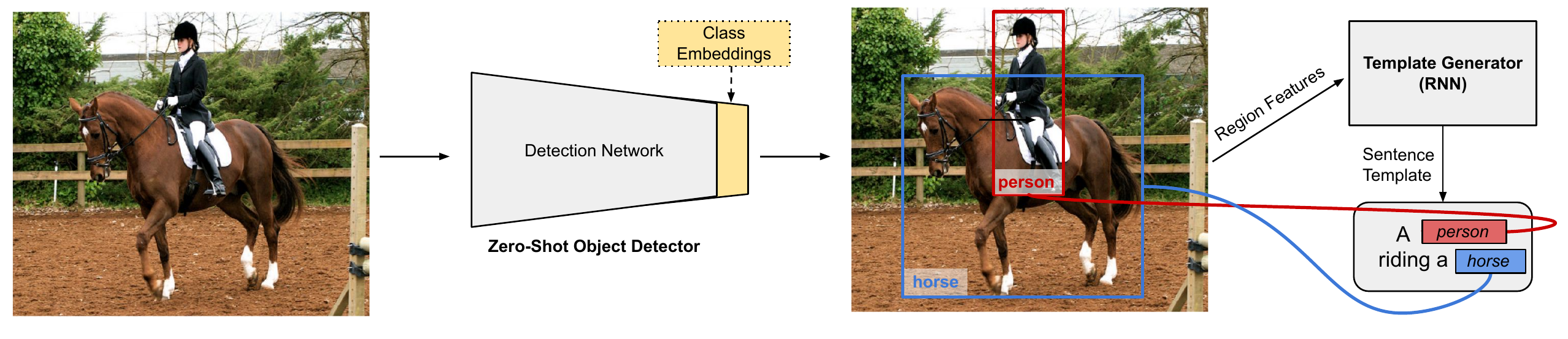}
\end{center}
    \caption{Our zero-shot captioning framework, which consists of two components: (i) zero-shot object detector, and (ii) image caption generator. For inference, input image is fed into the object detector, which provides the candidate objects and their representations.  The candidates are then given to the sentence generation RNN to obtain a sentence template, which consists of textual and visual words. The final sentence is obtained by setting the visual words according to the data obtained from the zero-shot detector.}
\label{main}
\end{figure*}

Following these observations, we argue that the true zero-shot captioning problem, where some of the test objects have no supervised
visual or textual examples, needs to be studied towards (i) developing semantically scalable captioning methods, and,
(ii) evaluating captioning approaches in a realistic setting where not all object classes have training examples.
To tackle this problem, we propose a novel ZSC approach that consists of a novel ZSD model and a template-based~\cite{lu2018neural} caption
generator defined over the ZSD outputs.  Our ZSD model uses the class-to-class similarities obtained over the distributed word
representations~\cite{mikolov2013distributed} between each target class and all training classes.
In order to address the generalized zero-shot detection (GZSD) problem, where test images contain a mixture seen and unseen
class instances, we propose a scaling scheme to make scores of seen and unseen classes more comparable, as illustrated
in Figure~\ref{main}.
For experimental evaluation, we use the MS-COCO dataset~\cite{lin2014microsoft} with train and test splits where the
model is trained over a strict subset of object classes. 

To sum up, 
this work aims to make a number conceptual, technical and experimental contributions in image captioning, which can be summarized as follows:
(i) we define a new paradigm for generating captions of unseen classes, (ii) we propose a novel ZSD approach that incorporates a probability scaling scheme for the generalized zero-shot object detection (GZSD) problem,
(iii) we evaluate several caption evaluation metrics and discuss their suitability for the zero-shot image captioning scenario.

\section{Related Work}
\label{sec:relwork}

\vspace{2mm}
\noindent \textbf{Zero-shot Learning.} Early work on ZSL focused on directly using attributes for transferring semantic information from seen to unseen classes~\cite{lampert09}. More recent work studies other knowledge transfer mediums, such as distributed word representations and
hierarchies \cite{akata2015evaluation}, visually consistent word representations~\cite{demirel2017attributes2classname}, 
as well as better knowledge transfer methods, such as synthesized classifiers~\cite{changpinyo2016synthesized}, semantic autoencoders~\cite{kodirov2017semantic}, hierarchy graphs~\cite{deng2014large}, diffusion regularization~\cite{long2018zero}, attribute regression~\cite{luo2018zero} and latent space encoding~\cite{yu2018zero}. There are also studies that aim to reduce the bias towards seen classes, \eg \cite{song2018transductive}.
More recently, data-generating models have been proposed for ZSL~\cite{zhu2018generative, wang2018zero, mishra2018generative, sariyildiz2019gradient}.

\vspace{2mm}

\noindent \textbf{Object Detection.} Most recent object detection methods can be categorized into the following two groups: (i) regression based approaches, and, (ii) region proposal based approaches. Regression based approaches work on detecting the candidate objects and their locations directly without any intermediate step~\cite{law2018cornernet, redmon2016yolo9000, redmon2016you, liu2016ssd, shen2017dsod, fu2017dssd}. Region proposal based approaches firstly generate region proposals, and then classify each region proposal into candidate classes with a confidence score~\cite{he2015spatial, dai2016r,  lin2018focal, lin2017feature, ren2015faster, sermanet2014overfeat, girshick2015fast}. Recently, zero-shot object detection methods are also emerging~\cite{rahman2018zero, demirel2018zero, bansal2018zero, rahman2018polarity}. Among these, Rahman \textit{et al.}~\cite{rahman2018zero} use semantic clustering loss for clustering similar classes. Demirel \textit{et al.}~\cite{demirel2018zero} propose a hybrid method to handle visually and semantically meaningful word vectors together. Bansal \textit{et al.}~\cite{bansal2018zero} propose a background-aware approach to learn a generalized zero-shot object detector. Rahman \textit{et al.}~\cite{rahman2018polarity} reshape embeddings so that visual features are well-aligned with related semantics. 
As an alternative to zero-shot detection, \cite{hoffman2014lsda, hoffman2015detector,redmon2016yolo9000} propose to
transfer knowledge from learned image classifiers to detectors, similar to weakly supervised learning~\cite{bilen2016weakly,cinbis2016weakly}. However, such approaches require labeled training images for all classes of interest, which significantly reduces their suitability to zero-shot captioning.

The approach closest to our ZSD method is the one proposed by Demirel \textit{et al.}~\cite{demirel2018zero}. Our approach differs by (i) leveraging 
class-to-class similarities measured in the word embedding space, instead of directly using the word embeddings, (ii) 
learning a class score scaling factor to improve generalized zero-shot detection performance. 

\vspace{2mm}

\noindent \textbf{Image Captioning.} Most recent image captioning are approaches are based on deep neural networks~\cite{mao2014deep, you2016image, xu2015show, kiros2014unifying, karpathy2015deep, kulkarni2013babytalk, lu2018neural}. Mainstream methods can be categorized as (i) template-based techniques~\cite{kulkarni2013babytalk, farhadi2010every, lu2018neural} and retrieval-based ones~\cite{hodosh2013framing, ordonez2011im2text,sun2015automatic,kiros2014unifying}. Template-based captioning approaches generate templates with empty slots, and fill those slots by using attributes or detected objects. Kulkarni \textit{et al.}~\cite{kulkarni2013babytalk} use conditional random fields (CRF) to push tight connections between the image content and sentence generation process before filling the empty slots. Farhadi \textit{et al.}~\cite{farhadi2010every} use triplets of scene elements for filling the empty slots in generated template. Lu \textit{et al.}~\cite{lu2018neural} use recurrent neural network to generate sentence templates for slot filling. Retrieval-based image captioning methods rely on retrieving captions from the set of training examples. More specifically, a set of training images similar to the test example are retrieved and the captioning is performed over their captions.

In this work, we aim to generate captions that can include classes that are not seen within in the supervised training set, where retrieval-based approaches are
not directly suitable. For this reason, we adopt a template-based approach that generates sentence templates and fills the visual word slots with the ZSD outputs.

Dense captioning~\cite{johnson2016densecap,yang2017dense,krishna2017dense} appear to be similar to the zero-shot image
captioning, but the focus is significantly different: while dense captioning aims to generate rich descriptions, in ZSC
our goal is to achieve captioning over novel object classes. Some captioning methods go beyond training with fully
supervised captioning data and allow learning with a captioning dataset that covers only some of the object classes plus
additional supervised examples for training object detectors and/or classifiers for all classes of
interest~\cite{anne2016deep, venugopalan2017captioning,yao2017incorporating, anderson2016guided, wu2018decoupled}. Since
these methods presume that all necessary visual information can be obtained from some pre-trained object recognition
model, we believe they cannot be seen as true zero-shot captioning approaches.

\section{Method}
\label{sec:method}
In this section, we describe the proposed ZSC model, and its components. It consists of a ZSD model that leverages the class-to-class similarities obtained over the distributed word representations, and a template-based image captioning model that generates sentence templates using a recurrent neural network.

In the rest of this section, we explain the details of the model components and then describe how we build the final ZSC model by using these components.

\subsection{Zero-Shot Object Detection} In zero-shot object detection, the goal is to learn a detection model
over the examples given for the seen classes, denoted by $Y_s$ such that the detector can recognize and localize the bounding boxes of
the instances of all classes $Y = Y_{s} \cup Y_{u}$, where $Y_{u}$ represents the set of unseen classes. 
For this purpose, we use the
YOLO~\cite{redmon2016you} architecture as our backend, and adapt it to the ZSD problem. 

In the
original YOLO approach, the loss function consists of three components: (i) the localization loss, which measures the error
between ground truth locations and predicted bounding boxes, (ii) the objectness loss, and, (iii) the recognition
loss.  The original
recognition loss $\ell_\text{cls}$ is the squared error of class conditional probabilities at each cell, in a grid of size $S \times S$ ($S=13$, by default):
\begin{equation}
\ell_{cls}(x) = \sum_{i=0}^{S^{2}} \mathds{1}_\text{object}^{i} \sum_{c \in Y_{s}}^{}(t(x,c,i)-f(x,c,i))^{2}
\label{eq:label_loss}
\end{equation}
where the target indicator mapping $t(x,c,i)$ is $1$ if the cell $i$ of image $x$ contains an instance of
class $c$ and otherwise $0$. $f(x,c,i)$ is the prediction score corresponding two the class $c$ and cell $i$. $\mathds{1}_\text{object}^{i}=1$ if an object instance appears in cell $i$, and otherwise $0$.

Following the approach in \cite{demirel2018zero}, we adapt the YOLO model to the ZSD problem by re-defining the prediction function
as a compatibility estimator between the {\em cell embeddings} and {\em class embeddings}, so that novel test classes can be 
predicted at the test time based purely on their class embeddings:
\begin{equation}
f(x,c,i) = \frac{\Omega(x,i)^{T}\Psi (c)}{\parallel \Omega(x,i) \parallel  \parallel \Psi(c) \parallel }
\label{eq:label_f}
\end{equation}
Here, $\Omega(x,i)$ denotes the cell embeddding obtained for the image cell $i$ for the input image $x$, and, $\Psi(c)$ represents
the class embedding for the class $c$. Unlike the approach in \cite{demirel2018zero}, which uses word embeddings (or attributes) of class names directly as the
class embeddings, we use the vector of class-to-class similarities in the word embedding space to obtain the class embeddings to obtain 
more compact and descriptive class embeddings.
More specifically, we define the class embedding for the class $c$ as the vector of similarities between $c$
and each reference training class $\bar{c}$:
\begin{equation}
\varphi(c)^{T}\varphi(\bar{c}) +1
\label{eq:label_sim}
\end{equation}
where $\varphi(c)$ denotes the distributed word representations~\cite{mikolov2013distributed} for class $c$. 

In this work, we need to obtain a generalized zero-shot detector, where both training and test classes appear at
the test images, as opposed to zero-shot detection as in \cite{demirel2018zero}. However, there can still be a significant
bias towards the seen classes, as the unseen test classes are not directly represented in the training set. We aim
to overcome this problem by introducing a scaling coefficient $\alpha$ for the unseen test classes as follows:
\begin{equation}
f(x,c,i) = \begin{cases}
\alpha f(x,c,i), & \text{ if } c \in \bar{Y_{s}} \\ 
f(x,c,i), & \text{ otherwise }
\end{cases}
\label{eq:label_bias}
\end{equation}
Here, $\bar{Y_{s}}$ represents the selected subset of classes from the training classes $Y_{s}$ to simulate the unseen
class scores.
In order to learn the coefficient $\alpha$, 
we first train the ZSD model over all training classes without $\alpha$, freeze the network, 
select
a subset of the training classes as unseen classes, and remove (\ie set to zero) the corresponding 
similarities in the class embeddings. Then, in this setting, we train only $\alpha$ to make
the scores among all training classes comparable. At test time, $\alpha$ is used as a scaling factor only for the unseen classes.

\subsection{Template-based Image Captioning}

We aim to generate accurate captions for images containing classes not seen during training. For this purpose, we use
a template-based captioning method which provides the sentence templates whose visual word slots are to be filled in
using the ZSD outputs. For this purpose, we use the slotted sentence template generation component of the Neural Baby Talk (NBT) approach~\cite{lu2018neural}.

The NBT method generates sentence templates which consist of the empty word slots by using a recurrent neural network. Moreover, NBT incorporates the pointer networks idea~\cite{vinyals2015pointer} to obtain a content-based attention mechanism over the
grounding regions. There are two word types in NBT method: textual and visual words. Textual words are not related
to any image region, therefore the model provides only dummy grounding for them. The template generation network uses the object
detection outputs to fill empty visual word slots.

In our approach, the ZSD model and the sentence template generation component of NBT is trained over the seen classes. The 
ZSD outputs over all classes are used as input to the NBT sentence generator at test time.

\section{Experiments}
\label{sec:experiments}

In this section, we present our experimental analysis for the ZSD component of our ZSC approach and the ZSC outputs
themselves.  For this purpose, we first define the COCO dataset splits that we use and the
the word embeddings used for computing the class-to-class similarities in
Section~\ref{sec:dataset}. We present the experimental results for the ZSD component in
Section~\ref{sec:detection}, and, those for the complete ZSC model in Section~\ref{sec:captioning}. Finally, we present ablative
studies in Section~\ref{sec:ablative}.

\subsection{Dataset}
\label{sec:dataset}

We use the COCO image captioning dataset~\cite{lin2014microsoft} in our experiments. As discussed in Section~\ref{sec:relwork},
there are several zero-shot captioning studies that assume that image-caption training example pairs are provided only for a subset of the object classes 
during training, while presuming the availability of visual training examples for all classes~\cite{anne2016deep, venugopalan2017captioning,yao2017incorporating, anderson2016guided, lu2018neural}. We refer to them as {\em partial zero-shot} methods and 
consider them as upper-bounds (golden baselines) for our true ZSC problem definition. In order to compare against these methods,
we use the same COCO dataset splits used by the respective publications. 

The details of the COCO splits are as follows: from the 80 classes, we select 8 of them as unseen classes ({\em bottle}, {\em bus}, {\em couch}, {\em microwave}, {\em pizza}, {\em tennis racket}, {\em suitcase}, and {\em zebra}). During training, we use the subset of the COCO training images that do not contain any instances of these unseen classes. For evaluation, we use the validation set split prepared by~\cite{anne2016deep}.
 
\paragraph{Word embeddings.} To obtain the class-to-class similarities required by our zero-shot detector, we use the $300$-dimensional word2vec
embeddings of class names \cite{mikolov2013efficient}. For the classes with multiple words, we use the
average word embeddings. For the NBT component of our ZSC approach, we use $300$-dimensional GloVe~\cite{pennington2014glove} word embeddings 
for template generation, as in the original NBT model.

\subsection{Zero-Shot Object Detection}
\label{sec:detection}

For training the ZSD component of our approach, we train the YOLO~\cite{redmon2016you} based network for $160$ epochs, with
the learning rate set to $0.001$ and a batch size of $32$. The detector training is conducted on COCO object detection dataset
using only the training examples of selected $72$ seen classes. Then, we select $64$ of these $72$ classes as seen and
remaining 8 of which as unseen, and learn the $\alpha$ scaling factor.

We evaluate the proposed zero-shot detection approach in ZSD and Generalized ZSD (GZSD) settings. For the ZSD evaluation, 
we use the COCO validation images that consist of only unseen object instances and by feeding only unseen classes as
target classes to the detector. The results are shown in the first column of Table~\ref{table:zsd_results}. On average, the
method yields $31.4\%$ mean average precision (mAP) score, which can be interpreted as a promising result considering the
difficulty of the zero-shot detection problem. 

For the GZSD evaluation, we use the COCO val5k split, which contains both seen and unseen class instances. We separately
compute mAP scores over the seen and unseen classes.  As the final scalar performance metric, we use the harmonic mean
(HM) of seen and unseen class mAP scores, following its use in generalized zero-shot classification
evaluation~\cite{xian2018zero}. The GZSD results without and with alpha scaling are presented in the last two rows 
of Table~\ref{table:zsd_results}, respectively. It can be seen that, with alpha-scaling, the unseen class mAP increases from
$0.3\%$ to $7.3\%$ mAP but the seen class mAP drops from $27.4\%$ to $19.2\%$ mAP. These results are as expected considering the 
fact that alpha-scaling acts as a prior that promotes the detection of unseen classes over the seen ones. Overall, the harmonic mean score
increases from $0.7\%$ to $10.6\%$ mAP using alpha-scaling, which suggests that alpha scaled detector is more suitable for being used 
as a component of the ZSC model.

\begin{table}
\centering
\resizebox{\columnwidth}{!}{%
\begin{tabular}{l|l|cccccccc|ccc}
Exp. Type        & Test & \multicolumn{1}{l}{bottle} & \multicolumn{1}{l}{bus} & \multicolumn{1}{l}{couch} & \multicolumn{1}{l}{microwave} & \multicolumn{1}{l}{pizza} & \multicolumn{1}{l}{racket} & \multicolumn{1}{l}{suitcase} & \multicolumn{1}{l|}{zebra} & \multicolumn{1}{l}{U-mAP(\%)} & \multicolumn{1}{l}{S-mAP(\%)} & \multicolumn{1}{l}{HM} \\ \hline
ZSD           & U    & 5.2                        & 53.3                    & 35.1                      & 23.9                          & 44.4                      & 36.4                       & 9.1                          & 43.7                       & 31.4                          & -                             & -                      \\ 
\hline
GZSD w/o $\alpha$ & S+U  & 0                          & 0                       & 2.7                       & 0                             & 0                         & 0                          & 0                            & 0                          & 0.3                           & 27.4                          & 0.7                    \\
GZSD           & S+U  & 0.8                        & 21.4                    & 4.9                       & 1.2                           & 4.8                       & 0.7                        & 9.1                          & 15.8                       & 7.3                           & 19.2                          & 10.6                  
\end{tabular}%
}
\caption{Our ZSD and Generalized ZSD (GZSD) results. The first row shows the ZSD results where the detector is evaluated over the images containing only unseen classes, without computing the seen class scores. The last two rows show the GZSD results where both seen and unseen class scores are computed on the full COCO val5k split. HM: Harmonic Mean.}
\label{table:zsd_results}
\end{table}

Finally, we note that these zero-shot detection results are not comparable to the previously reported COCO ZSD results of Bansal~\etal\cite{bansal2018zero}  
since the splits that we take from \cite{anne2016deep} (primarily for ZSC evaluation purposes) are different from those used in \cite{bansal2018zero}.

\subsection{Zero-Shot Image Captioning}
\label{sec:captioning}

For the ZSC experiments, we use our detection model in the GZSL setting. We obtain the candidate object regions by
setting the class confidence threshold of the ZSD detector to $0.5$, in order to retain only confident detections, as
needed by the template-based caption generator.  We prepare the same experimental setup described
in~\cite{lu2018neural}, and exclude the image-sentence pairs which consist of unseen classes.  To establish a fair
comparison, we use the NBT method~\cite{lu2018neural} for learning the language model. 

For comparison, we define the {\em NBT-baseline} by training a seen class only YOLO detector and feeding its outputs
to the captioning model. For evaluation, we use the METEOR~\cite{denkowski2014meteor}, SPICE~\cite{anderson2016spice} and Average F1 metrics
to evaluate the NBT baseline, the proposed method and the upper-bound partial ZSC methods, on the validation split of \cite{anne2016deep}.

\begin{table}[]
\centering
\resizebox{\columnwidth}{!}{%
\begin{tabular}{l|cccccccc|ccc}
Method     & \multicolumn{1}{l}{bottle} & \multicolumn{1}{l}{bus} & \multicolumn{1}{l}{couch} & \multicolumn{1}{l}{microwave} & \multicolumn{1}{l}{pizza} & \multicolumn{1}{l}{racket} & \multicolumn{1}{l}{suitcase} & \multicolumn{1}{l|}{zebra} & \multicolumn{1}{l}{Avg. F1} & \multicolumn{1}{l}{METEOR} & \multicolumn{1}{l}{SPICE} \\
\hline
NBT-baseline  & 0& 0& 0& 0& 0& 0& 0& 0& 0 & 18.2 & 12.7 \\
Our Method &2.4                       & 75.2                    & 26.6                      & 24.6                          & 29.8                      & 3.6                       & 0.6                         & 75.4                       & 29.8                       & 21.9                       & 14.2 \\
\hline
\multicolumn{12}{c}{Partial zero-shot captioning methods (upper-bound)}\\
\hline
DCC~\cite{anne2016deep}& 4.6                        & 29.8                    & 45.9                      & 28.1                          & 64.6                      & 52.2                       & 13.2                         & 79.9                       & 39.8                       & 21.0                       & 13.4                      \\
NOC~\cite{venugopalan2017captioning}& 17.8                       & 68.8                    & 25.6                      & 24.7                          & 69.3                      & 68.1                       & 39.9                         & 89.0                       & 49.1                       & 21.4                       & -                         \\
C-LSTM~\cite{yao2017incorporating}& 29.7                       & 74.4                    & 38.8                      & 27.8                          & 68.2                      & 70.3                       & 44.8                         & 91.4                       & 55.7                       & 23.0                       & -                         \\
Base+T4~\cite{anderson2016guided}& 16.3                       & 67.8                    & 48.2                      & 29.7                          & 77.2                      & 57.1                       & 49.9                         & 85.7                       & 54.0                       & 23.3                       & 15.9                      \\
NBT+G~\cite{lu2018neural}& 14.0                       & 74.8                    & 42.8                      & 63.7                          & 74.4                      & 19.0                       & 44.5                         & 92.0                       & 53.2                       & 23.9                       & 16.6                      \\
DNOC~\cite{wu2018decoupled}& 33.0                       & 77.0                    & 54.0                      & 46.6                          & 75.8                      & 33.0                       & 59.5                         & 84.6                       & 57.9                       & 21.6                       & -                                       
\end{tabular}%
}
\caption{Our ZSC results with comparison to upper-bound partial ZSC methods.}
\label{table:captioning}
\end{table}

In Table~\ref{table:captioning}, we see that the proposed approach obtains
$29.8\%$ average F1 score, which is computed by averaging per-class F1 scores. For the F1 score, a generated sentence is
counted as correct only if the sentence contains the name of the unseen class of interest in an image containing at
least one instance of that class. Here, an image is considered as a positive example only if it contains at least one
instance of the unseen object of interest.
We observe that the obtained F1 score of the proposed ZSC approach is promising but relatively lower than those of
the upper-bound partial ZSC methods, which is not a surprise considering that they are built upon supervised detectors
over all classes.

Compared to the upper-bound techniques in terms of METEOR and SPICE metrics,
our method obtains better results than DCC~\cite{anne2016deep} and NOC~\cite{venugopalan2017captioning} methods,
however lower than the other upper-bound methods, which benefit from using supervised object classification or detection models over all classes. 
Compared to the NBT-baseline, we observe significant improvements in terms of all metrics, which highlights the
importance of ZSD for the true ZSC problem.  

\begin{figure}
\begin{center}
\begin{tabular}{ccccccccc}
\bmvaHangBox{\includegraphics[width=3.8cm, height=3.4cm]{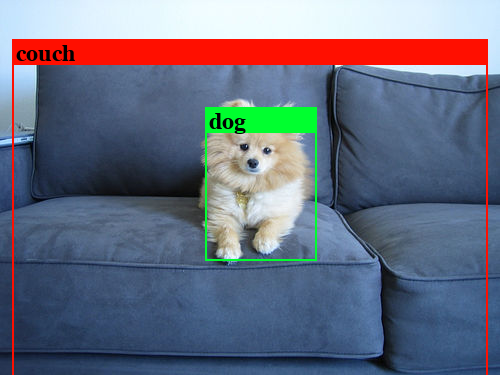}\\
A small white \textit{\textcolor{green}{dog}} sitting\\ on a \textbf{\textcolor{red}{couch}}.}
\bmvaHangBox{\includegraphics[width=3.8cm, height=3.4cm]{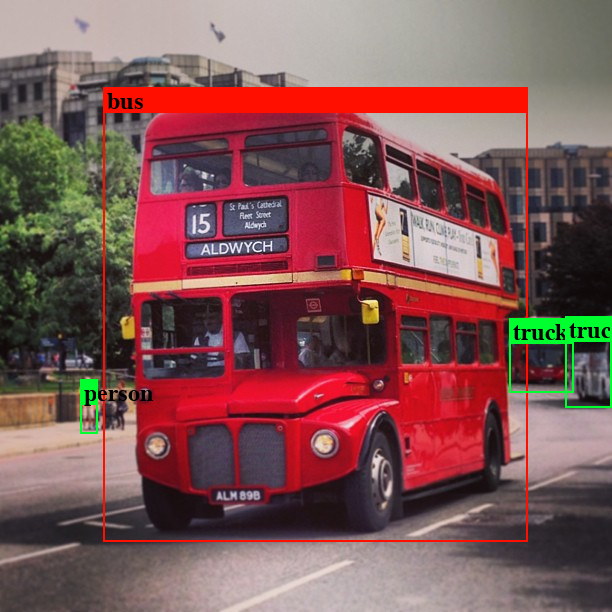}\\
A red \textbf{\textcolor{red}{bus}} is driving down\\ the street.}
\bmvaHangBox{\includegraphics[width=3.8cm, height=3.4cm]{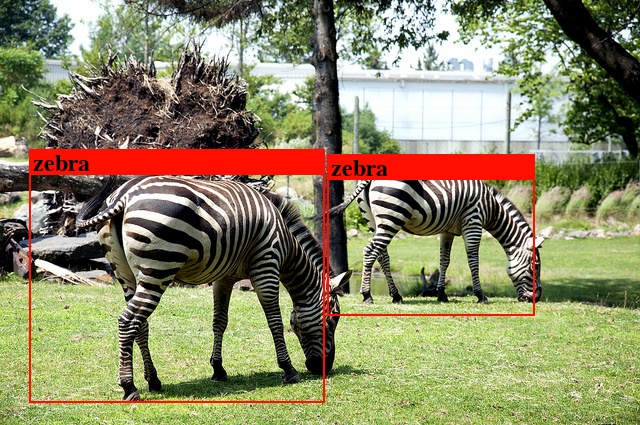}\\
A couple of \textbf{\textcolor{red}{zebra}} standing\\ in a field.}
\\
\bmvaHangBox{\includegraphics[width=3.8cm, height=3.4cm]{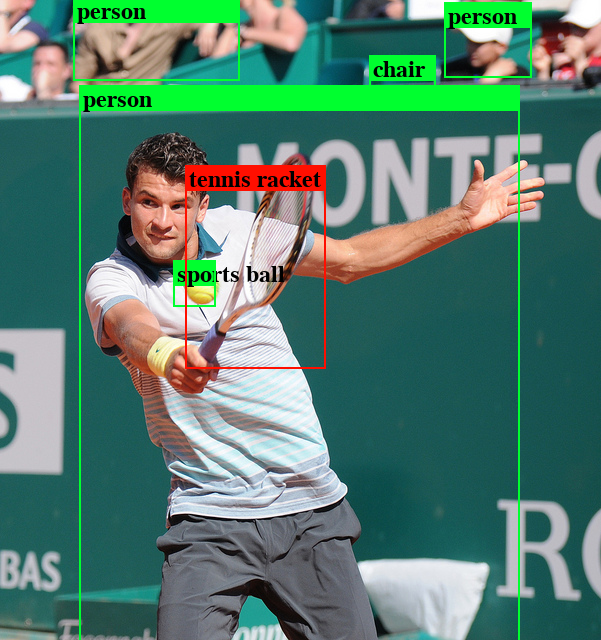}\\
A \textit{\textcolor{green}{tennis player}} is about\\ to hit a \textbf{\textcolor{red}{tennis racket}}.}
\bmvaHangBox{\includegraphics[width=3.8cm, height=3.4cm]{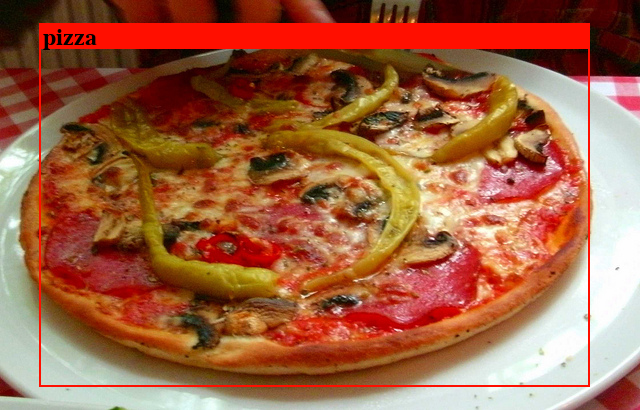}\\
A white plate topped with\\ a piece of \textbf{\textcolor{red}{pizza}}.}
\bmvaHangBox{\includegraphics[width=3.8cm, height=3.4cm]{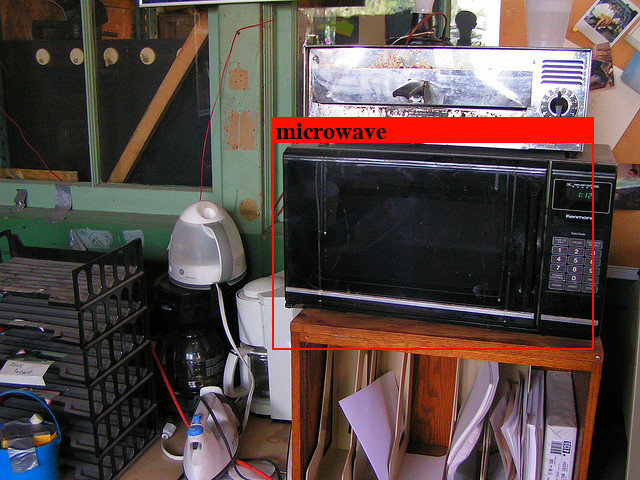}\\
A kitchen with a \textbf{\textcolor{red}{microwave}}\\ and a counter.}
\\
\bmvaHangBox{\includegraphics[width=3.8cm, height=3.4cm]{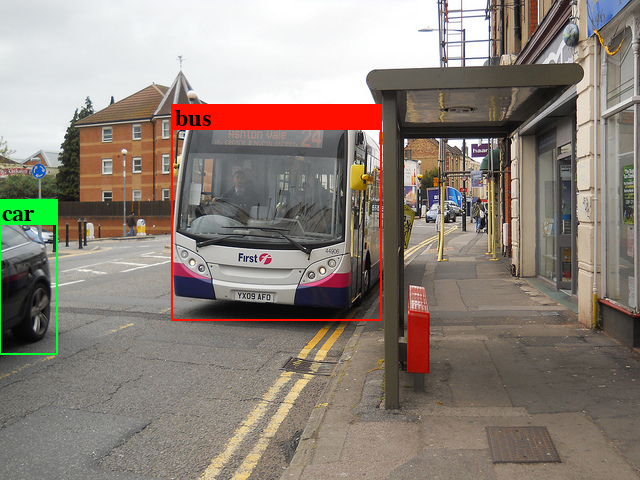}\\
A \textbf{\textcolor{red}{bus}} is parked on the\\ side of the street.}
\bmvaHangBox{\includegraphics[width=3.8cm, height=3.4cm]{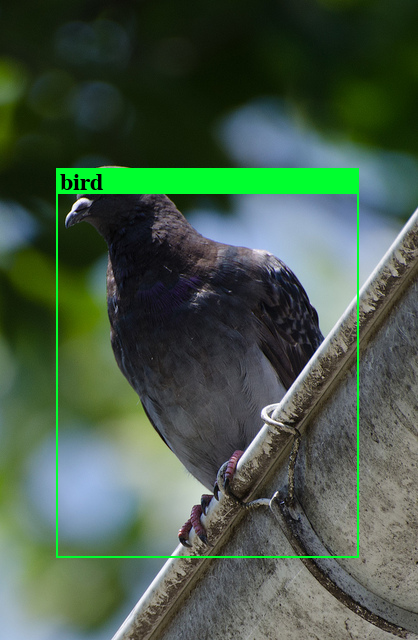}\\
A \textit{\textcolor{green}{bird}} sitting on top of a\\ metal pole.}
\bmvaHangBox{\includegraphics[width=3.8cm, height=3.4cm]{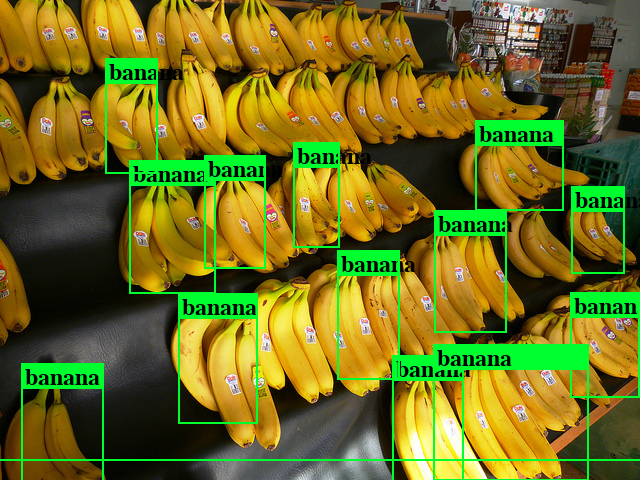}\\
A bunch of \textit{\textcolor{green}{banana}} that are\\ on a table.}
\end{tabular}
\end{center}
\caption{Image captioning results of images which consist of \textit{\textcolor{green}{seen}} and \textbf{\textcolor{red}{unseen}} classes.}
\label{fig:visual}
\end{figure}

Figure~\ref{fig:visual} gives qualitative ZSC results, where the seen ({\em italic} typeface) and unseen ({\bf bold}
typeface) class names are denoted.  Overall, these results are encouraging as the single final model shows (partial) ability to
generate captions with both seen and unseen classes, even in images where a group of seen and unseen class instances appear jointly.

\subsection{Ablative Studies}
\label{sec:ablative}

In this section, we measure the effect of the ZSD method on ZSC success in a more detailed manner. For this purpose, we compare our method with the NBT-baseline method, using the same split as in Section~\ref{sec:captioning}, by evaluating in terms of METEOR~\cite{denkowski2014meteor}, SPICE~\cite{anderson2016spice}, ROUGE-L~\cite{lin2004rouge}, BLEU~\cite{papineni2002bleu} metrics and the F1 score. 

We present the results in Table~\ref{table:ablative}. According to these results, we get the average F1 = 0 in NBT-baseline as expected, as it is unable to
caption with any unseen object instance. However, we observe that the results of the NBT-baseline and the proposed approach are very close to each other, especially in terms of BLUE and ROUGE-L metrics, despite the inability of the NBT-baseline to incorporate any unseen class in the captioning outputs. 
These results suggest that these metrics are not suitable for the ZSC scenario.

According to Table~\ref{table:ablative}, METEOR and SPICE metrics give slightly more indicative results for ZSC. We note that the METEOR metric uses $n$-gram based synonym matching and the SPICE metric uses scene-graph based synonym matching. On the other hand, remaining metrics use $n$-gram precision, $n$-gram recall, tf-idf weighted $n$-gram similarities without synonym matching. Therefore, we may preliminary conclude that using synonym matching contributes to obtaining more meaningful ZSC evaluations. 

\begin{figure}
\begin{tabular}{ccc}
\bmvaHangBox{\includegraphics[width=3.8cm, height=3.4cm]{images/captioning/COCO_val2014_000000119402.jpg}\\
$\blacklozenge$: A yellow and black \textbf{train}\\ traveling down the road.\\
$\bigstar$: A yellow and black \textbf{bus}\\ driving down a road.
}
\bmvaHangBox{\includegraphics[width=3.8cm, height=3.4cm]{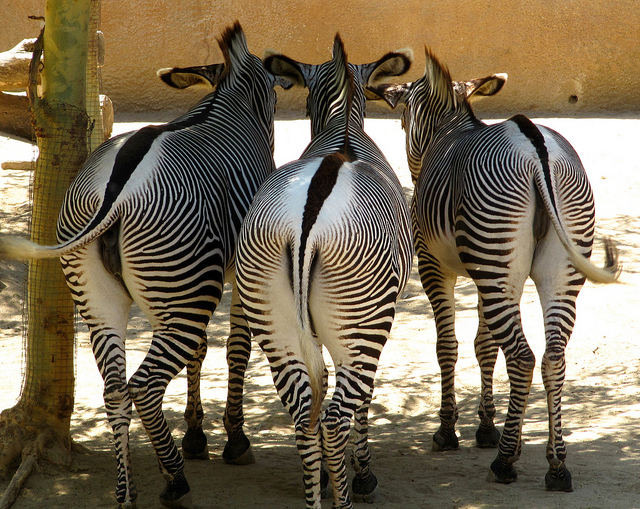}\\
$\blacklozenge$: A couple of \textbf{elephants}\\ standing next to each other.\\
$\bigstar$: A couple of \textbf{zebra} standing\\ next to each other.}
\bmvaHangBox{\includegraphics[width=3.8cm, height=3.4cm]{images/captioning/COCO_val2014_000000407162.jpg}\\
$\blacklozenge$: A piece of \textbf{cake} on a\\ white plate.\\
$\bigstar$: A piece of \textbf{pizza} on a\\ white plate.
}\\
\end{tabular}
\\
    \caption{Zero-shot image captioning results using the NBT-baseline ($\blacklozenge$) and the proposed approach ($\bigstar$). \textbf{Bold} typed words represent visual words given by the detectors.}
\label{fig:ablative}
\end{figure}

Finally, we show a qualitative comparison between the NBT-baseline and our method in Figure~\ref{fig:ablative}. These results show how the NBT baseline generates captions with somewhat similar but inaccurate objects due to its inability to handle unseen class instances. 
 
\begin{table}[]
\centering
\resizebox{\columnwidth}{!}{%
\begin{tabular}{lllllllll}
Method & BLEU1         & BLEU2         & BLEU3         & BLEU4         & METEOR        & SPICE         & ROUGE-L       & F1 Scores      \\ 
\midrule
NBT-baseline      & 65.8          & 47.5          & 34.1          & 24.4 & 18.2          & 12.7          & 48.2          & 0              \\
Our Method      & \textbf{67.0} & \textbf{48.3} & \textbf{34.5} & \textbf{24.5} & \textbf{21.9} & \textbf{14.2} & \textbf{48.9} & \textbf{29.8}
\end{tabular}%
}
\caption{Zero-shot captioning evaluation results  in terms of various performance metrics.}
\label{table:ablative}
\end{table}

\section{Conclusion} 

An important shortcoming of current image captioning methods aiming to learn with class-restricted
caption annotations is their inability to operate in a fully zero-shot learning setting.  These methods generate
captions for images which consist of classes not seen in captioning datasets, but they still assume the availability of a 
a fully supervised object classifier and/or detector over all classes of interest. To initiate research towards overcoming this
important weakness in captioning research, we define the zero-shot image captioning problem, and
propose a novel approach that consist of a zero-shot object detector and a detector-driven caption generator. 
Our experimental results based on the COCO dataset show that our method yields promising results towards achieving true ZSC
goals. In addition, the qualitative results show that proposed architecture is able to generate natural looking and visually grounded
captions in several challenging test cases.

\section{Acknowledgements}
This work was supported in part by the TUBITAK Grant 116E445.

\bibliography{paper}
\end{document}